\documentclass{article} % For LaTeX2e
\usepackage{iclr2022_conference,times}

\usepackage{amsmath,amsfonts,bm}

\def\eqref#1{equation~\ref{#1}}
\def\1{\bm{1}}

\def\mG{{\bm{G}}}

\def\mP{{\bm{P}}}

\def\mU{{\bm{U}}}
\def\mV{{\bm{V}}}

\def\mX{{\bm{X}}}

\def\mZ{{\bm{Z}}}

\DeclareMathAlphabet{\mathsfit}{\encodingdefault}{\sfdefault}{m}{sl}
\SetMathAlphabet{\mathsfit}{bold}{\encodingdefault}{\sfdefault}{bx}{n}

\newcommand{\R}{\mathbb{R}}

\usepackage{hyperref}
\usepackage{url}
\usepackage[pdftex]{graphicx}

\title{Attention-Free Keyword Spotting}

\author{Mashrur M. Morshed \& Ahmad Omar Ahsan\thanks{Equal contribution.} \\
Islamic University of Technology \\ 
Dhaka, Bangladesh \\
\texttt{\{mashrurmahmud,ahmadomar\}@iut-dhaka.edu}}
\iclrfinalcopy % Uncomment for camera-ready version, but NOT for submission.
\begin{document}

\maketitle

% Maybe add results of V1-35
\begin{abstract}
Till now, attention-based models have been used with great success in the keyword spotting problem domain. However, in light of recent advances in deep learning, the question arises whether self-attention is truly irreplaceable for recognizing speech keywords. We thus explore the usage of gated MLPs---previously shown to be alternatives to transformers in vision tasks---for the keyword spotting task. We provide a family of highly efficient MLP-based models for keyword spotting, with less than 0.5 million parameters. We show that our approach achieves competitive performance on Google Speech Commands V2-12 and V2-35 benchmarks with much fewer parameters than self-attention-based methods.

% We verify our approach on the Google Speech Commands V2-35 dataset and show that it is possible to obtain performance comparable to the state of the art without any apparent usage of self-attention.
\end{abstract}

\section{Introduction}

% Transformers \citep{vaswani2017attention} have been a highly disruptive innovation in deep learning. They were originally introduced in the domain of Natural Language Processing, and have since replaced classical recurrent neural networks as the default approach in many NLP tasks \citep{devlin2018bert,yang2019xlnet}. Not only have Transformers transformed the field of NLP, but they have also been proven to have competitive performance across several important problem domains—such as image classification \citep{dosovitskiy2020image,yuan2021tokens}, video classification \citep{neimark2021video}, object detection \citep{carion2020end}, automatic speech recognition \citep{liu2021tera,gulati2020conformer} and so on.

% Partly due to their remarkable success in vision tasks,  have lately been studied in the field of keyword spotting (KWS) \citep{berg21_interspeech,gong2021ast}, and have shown similar exceptional results. Recent research \citep{tolstikhin2021mlp,liu2021pay,melas2021you} shows that a core component of Transformers, self-attention, may not be necessary for achieving good performance in vision and language tasks. This finding necessitates a study on whether MLPs can be an alternative to self-attention, which has been a main focus of several recent state-of-the-art methods for the KWS problem. Our contributions can be summarized as follows:

Transformers \citep{vaswani2017attention} have shown remarkable success in Computer Vision tasks with the advent of the Vision Transformer (ViT) \citep{dosovitskiy2020image}. They have lately been studied in the field of keyword spotting (KWS). Several works \citep{berg21_interspeech,gong2021ast} have obtained exceptional results with ViT-like approaches on KWS.

 Recent research \citep{tolstikhin2021mlp,liu2021pay,melas2021you,touvron2021resmlp} shows that a core component of Transformers, self-attention, may not be necessary for achieving good performance in vision and language tasks. This finding necessitates a study on whether MLPs can be an alternative to self-attention, which has been a main focus of several state-of-the-art methods for the KWS problem. Our contributions can be summarized as follows:

\begin{enumerate}

\item We introduce the Keyword-MLP (KW-MLP), a memory-efficient, attention-free alternative to the Keyword Transformer (KWT) \citep{berg21_interspeech}. It achieves 97.63\% and 97.56\% accuracy on the Google Speech Commands V2-12 and V2-35 benchmarks \citep{warden2018speech} respectively---showing comparable performance to the KWT, while having much fewer parameters.

\item We distill smaller and shallower versions of KW-MLP, with the smallest having only 0.213 million parameters, and accuracies of 97.12\% and 97.17\% on Google Speech Commands V2-12 and V2-35 benchmarks respectively.

\end{enumerate}

\section{Related Work}

\subsection{Keyword Spotting}
Keyword spotting (KWS) deals with identifying some pre-specified speech keywords from an audio stream. As it is commonly used in always-on edge applications, KWS research often focuses on both accuracy and efficiency. While research in keyword spotting goes back to the 1960s \citep{teacher1967experimental}, most of the recent and relevant works have been focused on the Google Speech Commands dataset \citep{warden2018speech}, which has inspired numerous works and has rapidly grown to be the standard benchmark in this field. The dataset contains 1 second long audio clips, each containing an utterance of a word. Notably, there are two versions, V1 and V2, consisting of 30 and 35 keywords respectively. There is also a 12 keyword task for either version, where it is required to identify 10 keywords and two additional classes, `silence' and `unknown' (containing instances of the unused keywords).

Initial approaches to keyword spotting on the speech commands consisted of convolutional models \citep{warden2018speech}. \citet{majumdar2020matchboxnet}, \citet{mordido2021compressing} and \citet{zhang2017end} proposed lightweight CNN models with depth-wise separable convolutions. \citet{de2018neural} proposed using a convolutional recurrent model with attention, introducing the usage of attention in the KWS task. \citet{rybakov2020streaming} proposed a multi-headed, self-attention-based RNN (MHAtt-RNN). \citet{vygon2021learning} proposed an efficient representation learning method with triplet loss for KWS. While the state of the art in KWS at that time was the method of \citet{rybakov2020streaming}, it was empirically seen that triplet loss performed poorly with RNN-based models. The authors later obtained excellent results with ResNet \citep{tang2018deep} variants.

Recently, \citet{berg21_interspeech} and \citet{gong2021ast} proposed the Keyword Transformer (KWT) and Audio Spectrogram Transformer (AST) respectively. Both approaches are inspired by the success of the Vision Transformer (ViT) \citep{dosovitskiy2020image}, and show that patch-based transformer models with self-attention can obtain state of the art or comparable results on the keyword spotting task. A key difference between these two transformer approaches is that AST uses ImageNet \citep{imagenet} and Audioset \citep{audioset} pre-training. Furthermore, the best performing KWT models are trained with attention-based distillation method \citep{touvron2021training} using a teacher MHAtt-RNN \citep{rybakov2020streaming}.

\subsection{MLP-based Vision}
The Vision Transformer \citep{dosovitskiy2020image} has thus far shown the remarkable capability of Transformers on image and vision tasks. However, several recent works have questioned the necessity of self-attention in ViT. \citet{melas2021you} directly raises the question on the necessity of attention, and shows that the effectiveness of the Vision Transformer may be more related to the idea of the patch embedding rather than self-attention. \citet{tolstikhin2021mlp} proposed the MLP-Mixer, which performs token mixing and channel mixing on image patches/tokens, and shows competitive performance on the ImageNet benchmark. \citet{touvron2021resmlp} showed similarly good ImageNet results with ResMLP, a residual network with patch-wise and channel-wise linear layers. \citet{liu2021pay} proposed the gMLP, consisting of very simple channel projections and spatial projections with multiplicative gating---showing remarkable performance without any apparent use of self-attention.

\section{Keyword-MLP}
\label{sec:proposed-method}

Inputs to KW-MLP consist of mel-frequency cepstrum coefficients (MFCC). Let an arbitrary input MFCC be denoted as $\mX \in \R^{F\times{T}}$, where $F$ and $T$ are the frequency bins and time-steps respectively. We divide $\mX$ into patches of shape $F\times{1}$, getting a total of $T$ patches. Each patch is effectively a vector of mel-frequencies for a particular time-step. 

The $T$ patches are flattened, giving us $\mX_{0} \in \R^{T \times F}$. We then map $\mX_{0}$ to a higher dimension $d$, with a linear projection matrix $\mP_{0} \in \R^{F \times d}$, getting the frequency domain patch embeddings $\mX_{E}$.

\begin{equation}
    \mX_{E} = \mX_{0}\mP_{0}
\end{equation}

\begin{figure}[ht]
\centering
    \includegraphics[width=0.6\linewidth]{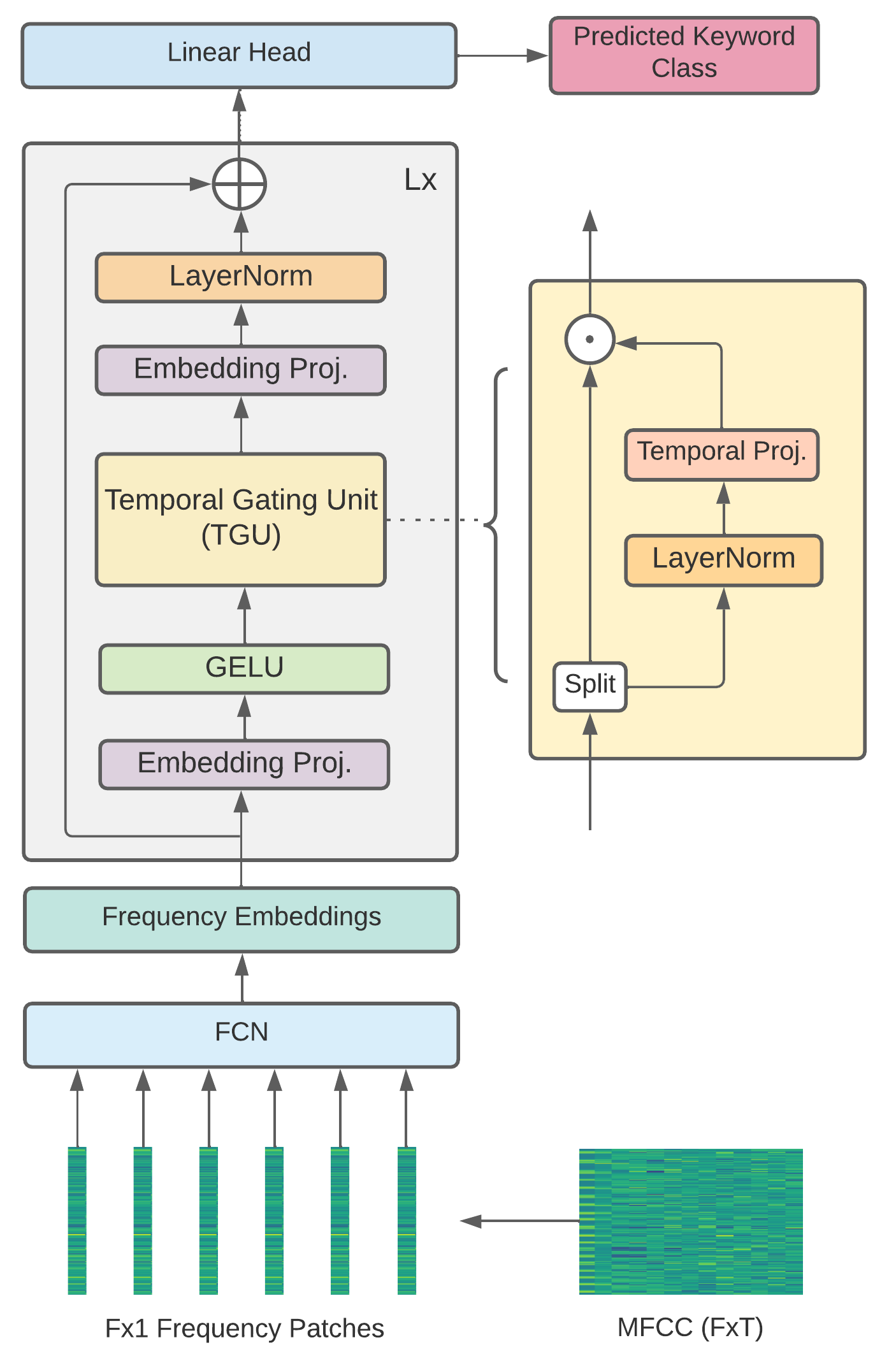}
\caption{The Keyword-MLP architecture, consisting of L blocks (equation \ref{eqn:gmlp}). Note that we move the LayerNorm to the end, before the skip-connection, different from \citet{liu2021pay} where norm is applied at the beginning. ($\oplus$ represents a residual skip-connection, while $\odot$ represents element-wise product.)}

\label{fig:kw-mlp}
\end{figure}

The obtained $\mX_{E}$ is passed through $L$ consecutive, identical gated-MLP (gMLP) blocks \citep{liu2021pay}. On a high level, we can summarize the gMLP blocks used in KW-MLP as a pair of projections across the embedding dimension separated by a projection across the temporal dimension. The block can also be formulated with the following set of equations (omitting bias and normalization for the sake of conciseness):

\begin{equation}
\begin{split}
\mZ  & =  \sigma (\mX_{in} \mU) \\
\tilde \mZ  =  g(\mZ) & = g([\mZ_{r} \mZ_{g}]) = \mZ_{r} \odot (\mZ_{g}^{T}\mG)^{T}  \\
\mX_{out}  & = \tilde \mZ\mV \oplus \mX_{in}
\end{split}
\label{eqn:gmlp}
\end{equation}

First, we use the matrix $\mU \in \R^{d\times{D}}$ to linearly project $\mX_{in}$ from the embedding dimension $d$ to the projection dimension $D$ (essentially a matmul operation). $\sigma$ represents the GELU activation function. $g(Z)$ represents the the Temporal Gating Unit (TGU) shown in Figure \ref{fig:kw-mlp}. The input to TGU, $\mZ$, is first split into $\mZ_{r}$ and $\mZ_{g} \in \R^{T\times{D/2}}$, the residual and the gate respectively. We use the matrix $\mG \in \R^{T\times{T}}$ to performs the linear projection across the temporal axis. This is followed by the linear gating---an element-wise multiplication with the residual $\mZ_{r}$. While the temporal projection operation can be implemented as passing $\mZ_{g}$ transposed through a \texttt{Dense(T, T)} layer, in practice, it can be implemented more efficiently by instead passing $\mZ_{g}$ through a \texttt{Conv1D(T, T, 1)} layer. $\tilde \mZ \in \R^{T\times{D/2}}$ is projected back to the embedding dimension $d$ with the matrix $\mV$ and then added with the skip-connected input $\mX_{in}$.

The original gMLP paper \citep{liu2021pay} applies LayerNorm \textit{before} the initial channel projection (analogous to embedding projections for images). We however find that applying norm \textit{after} the second embedding projection results in a notably faster and more optimal convergence. \citet{berg21_interspeech} also observe a similar phenomenon in their work.

The overall system is shown in Figure \ref{fig:kw-mlp}. In KW-MLP, we primarily use $L = 12$ (12 consecutive gMLP blocks), embedding dim $d = 64$, and projection dim $D = 256$. We also explore a group of smaller KW-MLP models with shallower depth, i.e. $L = 10, 8, 6$. The input MFCCs to the model are of shape $40\times98$, where 40 is the number of frequency bins, and 98 is the number of timesteps. All settings are also shown in Table \ref{tbl:settings}.

It is to be noted that the largest KW-MLP model has only 0.424M parameters, which is smaller than the smallest KWT variant (KWT-1 with 0.607M params) and much smaller than AST models (87M params). However, from Table \ref{tbl:accuracy} we can see that KW-MLP shows competitive accuracy with these models, particularly on the Speech Commands V2-35 benchmark.

\subsection{Knowledge Distillation}
In order to boost the accuracies of shallower KW-MLP variants, we use knowledge distillation (KD) \citep{hinton2015distilling}, using the KW-MLP model with $L = 12$ as the teacher model. We use an annealed KD approach \citep{jafari2021annealing} where the temperature parameter decreases every step following a cosine-annealing rule till it reaches 1. Other KD parameters, such as alpha, are shown in Table \ref{tbl:settings}. We also do not use label smoothing when training with KD, as soft targets are obtained from teacher predictions.

%%%%%%%%%%%%%%%%%%%%%%%%%%%%%
% RESULT TABLE
%%%%%%%%%%%%%%%%%%%%%%%%%%%%%
\begin{table}[ht]
	\centering
	\caption{Comparison of Model Parameters and Accuracy on Google Speech Commands V2-12 and V2-35 benchmarks \citep{warden2018speech}}
	\label{tbl:accuracy}
	\begin{tabular}	{l  l l l l }
		\hline
		Method    & Extra Knowledge & V2-12 & V2-35 & \# Params (M) \\
		\hline
		Att-RNN \quad [\citenum{de2018neural}]   & &  96.9 & 93.9  &  0.202\\
		Res-15 \quad [\citenum{vygon2021learning}]  & & 98.0 & 96.4 & 0.237\\
		MHAtt-RNN \quad [\citenum{rybakov2020streaming}]  & & 98.0 & 97.27 & 0.743\\
		\hline
		AST-S \quad [\citenum{gong2021ast}] & Pre. ImageNet  & & \textbf{98.11} & 87\\
		AST-P  \quad [\citenum{gong2021ast}] &  Pre. ImageNet  \& Audioset & & 97.88 & 87\\
		\hline
		KWT-3  \quad [\citenum{berg21_interspeech}] & KD with MHAtt-RNN & \textbf{98.56} & 97.69 & 5.361 \\
		KWT-2   \quad [\citenum{berg21_interspeech}] & KD with MHAtt-RNN & 98.43 & 97.74 & 2.394 \\
		KWT-1  \quad [\citenum{berg21_interspeech}]  & KD with MHAtt-RNN & 98.08 & 96.95 & 0.607 \\
		\hline
		KWT-3  \quad [\citenum{berg21_interspeech}] & & 98.54 & 97.51 & 5.361 \\
		KWT-2  \quad [\citenum{berg21_interspeech}]  & & 98.21 & 97.53 & 2.394 \\
		KWT-1  \quad [\citenum{berg21_interspeech}] & & 97.72 & 96.85 & 0.607 \\
		\hline
		KW-MLP  & & \textbf{97.63} & \textbf{97.56} & 0.424 \\
		KW-MLP$_{10}$   & &  97.38 & 97.35 & 0.353 \\
		KW-MLP$_{8}$   & & 97.28 & 97.26 & 0.283 \\
		KW-MLP$_{6}$   & & 97.03 & 97.07 & 0.213 \\
		\hline
		KW-MLP$_{10}$   & KD with KW-MLP & 97.30 &  97.49     & 0.353 \\
		KW-MLP$_{8}$   & KD with KW-MLP & 97.24  &  97.45     & 0.283 \\
		KW-MLP$_{6}$  & KD with KW-MLP & 97.12 & 97.17 & 0.213 \\
	\end{tabular}
\end{table}

% In the block, we move the LayerNorm to the end, just before the residual skip connection.

\section{Experimental Details}
We follow similar hyperparameters to \citet{rybakov2020streaming,berg21_interspeech}, with minor changes; all our hyperparameters and settings are shown in Table \ref{tbl:settings}. For training, we use a smaller batch-size of 256, and train for 140 epochs. No other augmentation apart from Spectral Augmentation \citep{park2019specaugment} is applied (to enable fast training). As an additional regularization method, each gMLP block has a survival probability of 0.9 (alternatively, a 0.1 probability to drop each block). We run experiments on Google Speech Commands V2-12 and V2-35 benchmarks, following the standard protocol described in \citet{warden2018speech}.

As seen from Table \ref{tbl:accuracy}, the largest KW-MLP model has only 424K parameters, which is much fewer than the KWT models, while having comparable accuracy. Furthermore, since we do not apply expensive run-time augmentations like resampling, time-shifting, adding background noise, mixup, etc. (used by \citep{rybakov2020streaming,berg21_interspeech,gong2021ast}), it is possible to train KW-MLP models in a very short time on free cloud compute such as the NVIDIA Tesla K80 or Tesla P100 provided by Google Colab and Kaggle. 

As a trade-off for fast training, a limitation of the KW-MLP experiments is that the effect of various augmentation methods have not been explored. This is more apparent in the V2-12 task, which contains much fewer training examples ($\sim37000$) than the V2-35 task ($\sim84000$). The KW-MLP model does not generalize as well here, as compared to V2-35.

%%%%%%%%%%%%%%%%%%%%%%%%%%%%%
% HPARAM TABLE
%%%%%%%%%%%%%%%%%%%%%%%%%%%%%
\begin{table}[ht]
	\centering
	\caption{Overview of Hyper-Parameters and Settings}
	\label{tbl:settings}
	\begin{tabular} {c c c c c c c c}
		\cline{1-2}  \cline{4-5} \cline{7-8}
		\multicolumn{2}{c}{Training} & & \multicolumn{2}{c}{Augmentation} & & \multicolumn{2}{c}{Model} \\
		\cline{1-2}  \cline{4-5} \cline{7-8}
		Epochs & 140 &  & \# Time Masks & 2 & &  \# Blocks, $L$ & 12 \\
		Batch Size & 256 & & Time Mask Width & [0, 25] & &  Input Shape & $40\times98$ \\
		Optimizer & AdamW & & \# Freq Masks & 2 & &  Patch Size & $40\times1$ \\
		Learning Rate & 0.001 & & Freq Mask Width & [0, 7] & & Dim, $d$ & 64 \\
		Warmup Epochs & 10 & & & & &  Dim Proj. & 256 \\
		Scheduling & Cosine &  & & & & \# Classes & 35 \\
		
		\cline{1-2}  \cline{4-5}  \cline{7-8}
		\multicolumn{2}{c}{Regularization} & & \multicolumn{2}{c}{Audio Processing} & & \multicolumn{2}{c}{KD}\\
		\cline{1-2}  \cline{4-5} \cline{7-8}
		Label Smoothing & 0.1 & & Sampling Rate & 16000 & & $\alpha$ & 0.9\\
		Weight Decay & 0.1 &  &  Window Length & 30 ms & & Init Temp & 5.0 \\
		Block Survival Prob. & 0.9 & &  Hop Length & 10 ms \\
		& & &  n\_mfcc & 40  \\

	\end{tabular}
\end{table}

\section{Temporal Projection Matrices}

We additionally visualize the weights of the temporal gating unit (the temporal projection matrix $\mG \in \R^{T\times{T}}$ in equation \ref{eqn:gmlp}). Interestingly, we can observe that our model learns weights which seem similar to diagonal, identity, or toeplitz matrices. This suggests that KW-MLP may partially learn a form of shift-invariance, which is necessary for the keyword spotting task. For instance, in a 1 second audio clip, a keyword can occur at different temporal positions; so the model needs to be invariant to temporal shift.

\begin{figure}[h]
\centering
    \includegraphics[width=0.6\linewidth]{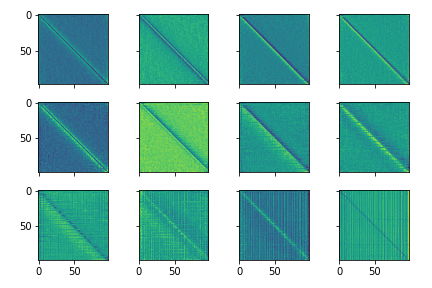}
\caption{Visualization of the temporal projection matrices, $\mG$, for each of the $L=12$ gMLP blocks of KW-MLP. The matrices are arranged in row major order.}
\label{fig:temp-matrix}
\end{figure}

\section{Conclusion}
The Keyword-MLP has shown itself to be an efficient solution to the keyword spotting task, and an alternative to self-attention-based methods. We hope that we provide an additional avenue of future research in audio and speech domains, particularly when resource-efficiency is concerned.

\bibliography{iclr2022_conference}

\begin{thebibliography}{23}
\providecommand{\natexlab}[1]{#1}
\providecommand{\url}[1]{\texttt{#1}}
\expandafter\ifx\csname urlstyle\endcsname\relax
  \providecommand{\doi}[1]{doi: #1}\else
  \providecommand{\doi}{doi: \begingroup \urlstyle{rm}\Url}\fi

\bibitem[Berg et~al.(2021)Berg, O’Connor, and Cruz]{berg21_interspeech}
Axel Berg, Mark O’Connor, and Miguel~Tairum Cruz.
\newblock {Keyword Transformer: A Self-Attention Model for Keyword Spotting}.
\newblock In \emph{Proc. Interspeech 2021}, pp.\  4249--4253, 2021.
\newblock \doi{10.21437/Interspeech.2021-1286}.

\bibitem[de~Andrade et~al.(2018)de~Andrade, Leo, Viana, and
  Bernkopf]{de2018neural}
Douglas~Coimbra de~Andrade, Sabato Leo, Martin Loesener Da~Silva Viana, and
  Christoph Bernkopf.
\newblock A neural attention model for speech command recognition.
\newblock \emph{arXiv preprint arXiv:1808.08929}, 2018.

\bibitem[Deng et~al.(2009)Deng, Dong, Socher, Li, Li, and Fei-Fei]{imagenet}
Jia Deng, Wei Dong, Richard Socher, Li-Jia Li, Kai Li, and Li~Fei-Fei.
\newblock Imagenet: A large-scale hierarchical image database.
\newblock In \emph{2009 IEEE Conference on Computer Vision and Pattern
  Recognition}, pp.\  248--255, 2009.
\newblock \doi{10.1109/CVPR.2009.5206848}.

\bibitem[Dosovitskiy et~al.(2020)Dosovitskiy, Beyer, Kolesnikov, Weissenborn,
  Zhai, Unterthiner, Dehghani, Minderer, Heigold, Gelly,
  et~al.]{dosovitskiy2020image}
Alexey Dosovitskiy, Lucas Beyer, Alexander Kolesnikov, Dirk Weissenborn,
  Xiaohua Zhai, Thomas Unterthiner, Mostafa Dehghani, Matthias Minderer, Georg
  Heigold, Sylvain Gelly, et~al.
\newblock An image is worth 16x16 words: Transformers for image recognition at
  scale.
\newblock \emph{arXiv preprint arXiv:2010.11929}, 2020.

\bibitem[Gemmeke et~al.(2017)Gemmeke, Ellis, Freedman, Jansen, Lawrence, Moore,
  Plakal, and Ritter]{audioset}
Jort~F. Gemmeke, Daniel P.~W. Ellis, Dylan Freedman, Aren Jansen, Wade
  Lawrence, R.~Channing Moore, Manoj Plakal, and Marvin Ritter.
\newblock Audio set: An ontology and human-labeled dataset for audio events.
\newblock In \emph{Proc. IEEE ICASSP 2017}, New Orleans, LA, 2017.

\bibitem[Gong et~al.(2021)Gong, Chung, and Glass]{gong2021ast}
Yuan Gong, Yu-An Chung, and James Glass.
\newblock {AST: Audio Spectrogram Transformer}.
\newblock In \emph{Proc. Interspeech 2021}, pp.\  571--575, 2021.
\newblock \doi{10.21437/Interspeech.2021-698}.

\bibitem[Hinton et~al.(2015)Hinton, Vinyals, and Dean]{hinton2015distilling}
Geoffrey Hinton, Oriol Vinyals, and Jeff Dean.
\newblock Distilling the knowledge in a neural network.
\newblock \emph{arXiv preprint arXiv:1503.02531}, 2015.

\bibitem[Jafari et~al.(2021)Jafari, Rezagholizadeh, Sharma, and
  Ghodsi]{jafari2021annealing}
Aref Jafari, Mehdi Rezagholizadeh, Pranav Sharma, and Ali Ghodsi.
\newblock Annealing knowledge distillation.
\newblock \emph{arXiv preprint arXiv:2104.07163}, 2021.

\bibitem[Liu et~al.(2021)Liu, Dai, So, and Le]{liu2021pay}
Hanxiao Liu, Zihang Dai, David~R So, and Quoc~V Le.
\newblock Pay attention to mlps.
\newblock \emph{arXiv preprint arXiv:2105.08050}, 2021.

\bibitem[Majumdar \& Ginsburg(2020)Majumdar and
  Ginsburg]{majumdar2020matchboxnet}
Somshubra Majumdar and Boris Ginsburg.
\newblock Matchboxnet: 1d time-channel separable convolutional neural network
  architecture for speech commands recognition.
\newblock \emph{arXiv preprint arXiv:2004.08531}, 2020.

\bibitem[Melas-Kyriazi(2021)]{melas2021you}
Luke Melas-Kyriazi.
\newblock Do you even need attention? a stack of feed-forward layers does
  surprisingly well on imagenet.
\newblock \emph{arXiv preprint arXiv:2105.02723}, 2021.

\bibitem[Mordido et~al.(2021)Mordido, Van~Keirsbilck, and
  Keller]{mordido2021compressing}
Gon{\c{c}}alo Mordido, Matthijs Van~Keirsbilck, and Alexander Keller.
\newblock Compressing 1d time-channel separable convolutions using sparse
  random ternary matrices.
\newblock \emph{arXiv preprint arXiv:2103.17142}, 2021.

\bibitem[Park et~al.(2019)Park, Chan, Zhang, Chiu, Zoph, Cubuk, and
  Le]{park2019specaugment}
Daniel~S Park, William Chan, Yu~Zhang, Chung-Cheng Chiu, Barret Zoph, Ekin~D
  Cubuk, and Quoc~V Le.
\newblock Specaugment: A simple data augmentation method for automatic speech
  recognition.
\newblock \emph{arXiv preprint arXiv:1904.08779}, 2019.

\bibitem[Rybakov et~al.(2020)Rybakov, Kononenko, Subrahmanya, Visontai, and
  Laurenzo]{rybakov2020streaming}
Oleg Rybakov, Natasha Kononenko, Niranjan Subrahmanya, Mirko Visontai, and
  Stella Laurenzo.
\newblock Streaming keyword spotting on mobile devices.
\newblock \emph{arXiv preprint arXiv:2005.06720}, 2020.

\bibitem[Tang \& Lin(2018)Tang and Lin]{tang2018deep}
Raphael Tang and Jimmy Lin.
\newblock Deep residual learning for small-footprint keyword spotting.
\newblock In \emph{2018 IEEE International Conference on Acoustics, Speech and
  Signal Processing (ICASSP)}, pp.\  5484--5488. IEEE, 2018.

\bibitem[Teacher et~al.(1967)Teacher, Kellett, and
  Focht]{teacher1967experimental}
C~Teacher, H~Kellett, and L~Focht.
\newblock Experimental, limited vocabulary, speech recognizer.
\newblock \emph{IEEE Transactions on Audio and Electroacoustics}, 15\penalty0
  (3):\penalty0 127--130, 1967.

\bibitem[Tolstikhin et~al.(2021)Tolstikhin, Houlsby, Kolesnikov, Beyer, Zhai,
  Unterthiner, Yung, Keysers, Uszkoreit, Lucic, et~al.]{tolstikhin2021mlp}
Ilya Tolstikhin, Neil Houlsby, Alexander Kolesnikov, Lucas Beyer, Xiaohua Zhai,
  Thomas Unterthiner, Jessica Yung, Daniel Keysers, Jakob Uszkoreit, Mario
  Lucic, et~al.
\newblock Mlp-mixer: An all-mlp architecture for vision.
\newblock \emph{arXiv preprint arXiv:2105.01601}, 2021.

\bibitem[Touvron et~al.(2021{\natexlab{a}})Touvron, Bojanowski, Caron, Cord,
  El-Nouby, Grave, Izacard, Joulin, Synnaeve, Verbeek,
  et~al.]{touvron2021resmlp}
Hugo Touvron, Piotr Bojanowski, Mathilde Caron, Matthieu Cord, Alaaeldin
  El-Nouby, Edouard Grave, Gautier Izacard, Armand Joulin, Gabriel Synnaeve,
  Jakob Verbeek, et~al.
\newblock Resmlp: Feedforward networks for image classification with
  data-efficient training.
\newblock \emph{arXiv preprint arXiv:2105.03404}, 2021{\natexlab{a}}.

\bibitem[Touvron et~al.(2021{\natexlab{b}})Touvron, Cord, Douze, Massa,
  Sablayrolles, and J{\'e}gou]{touvron2021training}
Hugo Touvron, Matthieu Cord, Matthijs Douze, Francisco Massa, Alexandre
  Sablayrolles, and Herv{\'e} J{\'e}gou.
\newblock Training data-efficient image transformers \& distillation through
  attention.
\newblock In \emph{International Conference on Machine Learning}, pp.\
  10347--10357. PMLR, 2021{\natexlab{b}}.

\bibitem[Vaswani et~al.(2017)Vaswani, Shazeer, Parmar, Uszkoreit, Jones, Gomez,
  Kaiser, and Polosukhin]{vaswani2017attention}
Ashish Vaswani, Noam Shazeer, Niki Parmar, Jakob Uszkoreit, Llion Jones,
  Aidan~N Gomez, {\L}ukasz Kaiser, and Illia Polosukhin.
\newblock Attention is all you need.
\newblock In \emph{Advances in neural information processing systems}, pp.\
  5998--6008, 2017.

\bibitem[Vygon \& Mikhaylovskiy(2021)Vygon and
  Mikhaylovskiy]{vygon2021learning}
Roman Vygon and Nikolay Mikhaylovskiy.
\newblock Learning efficient representations for keyword spotting with triplet
  loss.
\newblock \emph{arXiv preprint arXiv:2101.04792}, 2021.

\bibitem[Warden(2018)]{warden2018speech}
Pete Warden.
\newblock Speech commands: A dataset for limited-vocabulary speech recognition.
\newblock \emph{arXiv preprint arXiv:1804.03209}, 2018.

\bibitem[Zhang \& Koishida(2017)Zhang and Koishida]{zhang2017end}
Chunlei Zhang and Kazuhito Koishida.
\newblock End-to-end text-independent speaker verification with triplet loss on
  short utterances.
\newblock In \emph{Interspeech}, pp.\  1487--1491, 2017.

\end{thebibliography}
\bibliographystyle{iclr2022_conference}

% \appendix
% \section{Appendix}
% You may include other additional sections here.

\end{document}